\pdfoutput=1

\documentclass[11pt]{article}

\usepackage[final]{acl}

\usepackage{times}
\usepackage{latexsym}

\usepackage[T1]{fontenc}

\usepackage[utf8]{inputenc}

\usepackage{microtype}

\usepackage{pgfplots}
\usepackage{algorithm}
\usepackage{algpseudocode}
\usepackage{amssymb}
\usepackage{arydshln}
\usepackage{tcolorbox}

\usepackage{inconsolata}
\usepackage{amsmath} 
\usepackage{amsthm}
\usepackage{multirow}
\usepackage{booktabs}
\usepackage{hyperref}       
\usepackage{url}            
\usepackage{booktabs}       
\usepackage{amsfonts}       
\usepackage{nicefrac}       
\usepackage{microtype}      
\usepackage{lipsum}		
\usepackage{graphicx}
\usepackage{natbib}
\usepackage{doi}
\title{Improving Attributed Text Generation of Large Language Models via Preference Learning}

\author{
    \begin{tabular}{c}
    Dongfang Li, Zetian Sun, Baotian Hu, Zhenyu Liu, Xinshuo Hu, Xuebo Liu, Min Zhang \vspace{.5mm} \\
    \end{tabular}
    \\ \vspace{.5mm}
    \begin{tabular}{c}
    Harbin Institute of Technology (Shenzhen), Shenzhen, China \\
    \texttt{\{lidongfang,hubaotian,zhangmin2021\}@hit.edu.cn} \\ 
    \end{tabular}
     \\ \vspace{.5mm}
}

\begin{document}
\maketitle
\begin{abstract}


Large language models have been widely adopted in natural language processing, yet they face the challenge of generating unreliable content. Recent works aim to reduce misinformation and hallucinations by resorting to attribution as a means to provide evidence (i.e., citations).   However, current attribution methods usually focus on the retrieval stage and automatic evaluation that neglect mirroring the citation mechanisms in human scholarly writing to bolster credibility.
In this paper, we address these challenges by modeling the attribution task as preference learning and introducing an Automatic Preference Optimization (APO) framework. First, we create a curated collection for post-training with 6,330 examples by collecting and filtering from existing datasets. Second, considering the high cost of labeling preference data, we further propose an automatic method to synthesize attribution preference data resulting in 95,263 pairs. Moreover, inspired by the human citation process, we further propose a progressive preference optimization method by leveraging fine-grained information. 
Extensive experiments on three datasets (i.e., ASQA, StrategyQA, and ELI5) demonstrate that APO achieves state-of-the-art citation F1 with higher answer quality.~\footnote{We will release the code upon publication.}

\end{abstract}

\section{Introduction}
Large Language Models (LLMs) have demonstrated emergent abilities and have gained widespread application in Natural Language Processing (NLP)~\cite{gpt3,DBLP:journals/tmlr/WeiTBRZBYBZMCHVLDF22,chatgpt,DBLP:journals/corr/abs-2312-11805}. For example, LLMs have shown remarkable in-context learning capabilities across a variety of domains and tasks~\cite{dong2022survey}. Although LLMs have been widely adopted, a prominent issue is that they produce hallucinations in certain situations~\cite{Ye2023CognitiveMA,Zhang2023SirensSI}. In other words, they generate information that sounds plausible but is nonfactual, thereby limiting their applicability in the real world. To mitigate hallucinations, researchers have resorted to grounding statements in responses generated by LLMs to supported evidence, either by providing rationales or by adding citations to the statements~\cite{DBLP:journals/corr/abs-2311-03731,DBLP:conf/emnlp/LiuZL23}.

Recent works have utilized external knowledge sources such as retrieved documents and knowledge graphs for attribution~\cite{DBLP:conf/emnlp/0001PCKW21,li2023verifiable}. Generally, these works are divided into two types: 1) the model generates an answer with citations based on the retrieved documents~\cite{DBLP:journals/corr/abs-2311-07838}; 2) an answer is first generated, then modified again to add attribution references by retrieving with query and initial answer~\cite{DBLP:conf/acl/GaoDPCCFZLLJG23}. However, these works focus mainly on the retrieval stage~\cite{ye2023effective} and the evaluation process~\cite{DBLP:journals/corr/abs-2305-06311}. Considering the selection of the model's desired responses and behavior from its very broad knowledge and capabilities, it is more necessary to optimize the generation process, not only reducing the hallucination of the original answer but also avoiding the hallucination of the attribution process.
On the other hand, fine-tuning LLMs after pre-training can also significantly improve performance for users' downstream tasks. First, given positive examples of correct behavior, supervised fine-tuning can be performed using standard likelihood-based training. Secondly, given positive and negative examples (binary feedback or pairwise feedback), methods such as unlikelihood training on negative examples~\cite{DBLP:conf/iclr/WelleckKRDCW20} or RLHF-PPO~\cite{DBLP:journals/corr/abs-1909-08593} can be used for learning. However, these methods usually suffer from expensive data collection process, reward model training, sparse reward and text degeneration problems, making them difficult to use in practical applications~\cite{DBLP:journals/corr/abs-2310-12036}.

\begin{figure}[t]
    \centering
    \includegraphics[width=\linewidth]{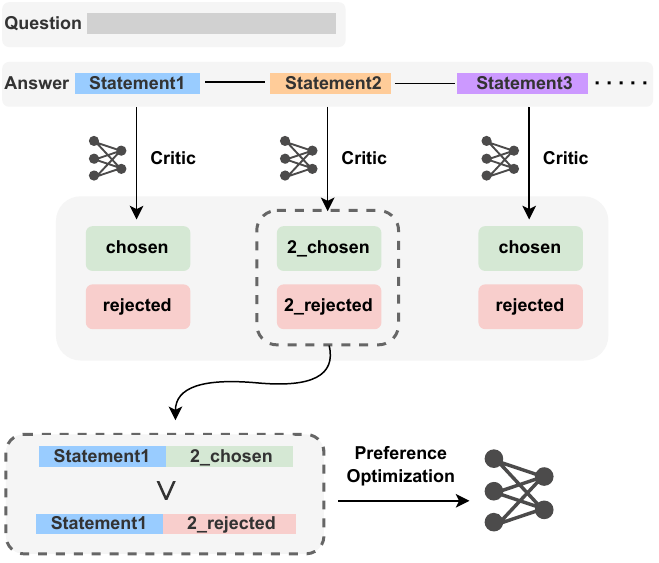}
    \caption{A brief overview of our APO framework. We show APO in more detail in Figure~\ref{fig:framework}.}
    \vspace{-4mm}
    \label{fig:intro}
\end{figure}

In this paper, inspired by the citation mechanisms in human scholarly writing~\cite{brooks1986evidence,teplitskiy2022status}, we address these challenges by conceptualizing the attribution task for LLMs as preference learning and proposing an Automatic Preference Optimization (APO) framework, as shown in Figure~\ref{fig:intro}. Initially, we assemble a curated dataset comprising 6,330 examples sourced and refined from existing datasets for post-training. This step makes the LLMs know the basic format and requirements of attribution.
Considering the substantial cost and extremely time-consuming of preference pair annotations, we thus introduce an automated approach to generate attribution preference data, yielding 95,263 pairs. Furthermore, drawing inspiration from the human process of citation and direct preference optimization~\cite{DBLP:journals/corr/abs-2305-18290}, we propose a progressive preference optimization method with experience replay bypassing the need for explicit reward modeling or reinforcement learning. 
We conduct the extensive experiment on three datasets (i.e., ASQA, StrategyQA, and ELI5). The experiment results demonstrate that APO surpasses compared baselines across all datasets with improved citation F1 along with higher response quality. Our contributions are summarized as follows:
\begin{itemize}
    \item To the best of our knowledge, we are the first to apply preference learning for attribution tasks. We also show that our method can be applied under synthesized preference scenarios.
    \item We establish a full data collection pipeline for attribution tasks and will open-source our all authorized data after publication for future research. 
    \item We propose a progressive preference optimization method to alleviate the sparse reward problem by leveraging fine-grained information. We further benchmark existing direct preference optimization methods and provide insights for attribution tasks. 
   
\end{itemize} 


\section{Related Work}


\subsection{Text Generation for Verification}
Prior works have studied methods and evaluations for verification that identify supporting sources for model outputs. For instance, ~\citet{DBLP:journals/corr/abs-2112-12870} introduce the concept of Attributable to Identified Sources (AIS) which transforms model outputs into standalone, interpretable propositions. The response \textit{s} can be attributed to a source \textit{P} if they meet the intuitive criterion ``\textit{According to P, s}''.~\citet{DBLP:journals/corr/abs-2212-08037} adapt the AIS framework for QA scenarios. Further,~\citet{DBLP:conf/emnlp/GaoYYC23} extrapolate AIS to evaluate  generated text of LLMs with citations. Additionally, several works focus on building and using automated AIS evaluations ~\cite{DBLP:conf/naacl/HonovichAHTKCSS22,DBLP:conf/acl/GaoDPCCFZLLJG23,DBLP:conf/emnlp/LiuZL23}.  For a comprehensive overview, please refer to~\citet{DBLP:journals/corr/abs-2311-03731}. In contrast to existing approaches, our work broadens the scope of attribution beyond just verifiable text generation and devises a methodology to enhance these attributions which frames it as a preference learning problem.

\subsection{Preference Optimization Methods}
Preference Optimization (PO) methods significantly improve generate quality to align with human values~\cite{DBLP:conf/nips/ChristianoLBMLA17,DBLP:journals/corr/abs-1909-08593,DBLP:conf/nips/StiennonO0ZLVRA20,DBLP:journals/corr/abs-2204-05862}. It usually first collects pairs of generations under the same context and a pairwise human preference to indicate which generation is better. Then the PO is used to optimize generating policy to generate better candidates from the pair. 
For example, Reinforcement Learning from Human Feedback (RLHF) is a model-based algorithm to optimize preference learning~\cite{DBLP:conf/nips/Ouyang0JAWMZASR22}.
However, the RLHF process is complex, time-consuming, and unstable. The direct PO uses an off-policy algorithm to directly optimize the generating policy, eliminating the need for a reward model~\cite{DBLP:journals/corr/abs-2305-18290,an2023dppo,OPPO,SLiC-HF}. These approach are more data-efficient and stable. 
For example, DPO uses the Bradley-Terry model ~\cite{bradley1952rank} and log-loss, which can lead to over-fitting to the preference data, especially when preference is deterministic and ignores the KL-regularization term. 
The IPO algorithm~\cite{DBLP:journals/corr/abs-2310-12036} addresses this issue by using a root-finding MSE loss to solve the problem of ignoring KL-regularization when preference is deterministic. However, these methods fail to fully account for more fine-grained preferences and that is exactly what we want to do.




\section{Preliminary}
The main pipeline of preference learning usually consists of: 1) pretraining and Supervised Fine-Tuning (SFT), where SFT is not a must; 2) preference data collection;
3) preference optimization.

\paragraph{Pretraining and SFT Phase}
Preference learning typically starts with a pretrained LLMs or LLMs fine-tuned on high-quality data using maximum likelihood estimation.
The final policy $\pi_{\text{ref}}$ after this phase is represented as
\begin{equation} \label{ref policy}
  \small \pi_{\text{ref}}\approx\arg\max_{\pi}\mathbb{E}_{x,y\sim\mathcal{D}_\text{ref}}\log \pi(x)\log(y|x),
\end{equation}
where $\mathcal{D}_\text{ref}$ denotes the training data distribution.

\paragraph{Preference Data Collection Phase}
After pretraining and SFT phase, $\pi_{\text{ref}}$ is prompted by context $x$, 
and generate two responses $y_w, y_l \sim \pi_\textnormal{ref}(\cdot|x)$. Then $x,y_w,y_l$ is labeled by humans to judge which response is preferred and denote $y_w\succ y_l | x$ if $y_w$ is preferred,
and $y_l\succ y_w|x$ if $y_l$ is preferred. We define a new symbol $I=\mathbb{I}[y_w\succ y_l|x]$, and all <$x,y_w,y_l,I$> consist the preference dataset $\mathcal{D}^p$:
\begin{equation}
  \langle x,y_w,y_l,I\rangle\sim\mathcal{D}^p.
\end{equation}

\paragraph{Preference Optimization Phase}
In the final phase, the prevailing method uses reinforcement learning algorithm to learn an explicit or implicit
reward from the preference data, and then using on-policy or off-policy policy gradient algorithm to maximize the reward.
Recently, some methods have derived the optimal policy using reward maximization under KL-regularization and also 
derive a loss with optimal policy as its solution,
then learn the optimal policy by minimizing the derived loss on empirical dataset.

\paragraph{Reinforcement Learning from Human Feedback (RLHF)} The RLHF uses standard two-phase reward model-based reinforcement learning to maximize the reward. It contains two steps:
1) reward estimation from preference data 2) reward maximization using PPO algorithm.
It aims to maximize reward with a KL constraint on the reference model $\pi_{\text{ref}}$ (inputs $x$ omitted):
\begin{equation}
\small
\pi^{*} = \arg\max_{\pi} \mathbb{E}_{
y \sim \pi} \left[ r(y) - \beta \log \frac{\pi(y)}{\pi_{\text{ref}}(y)} \right],
\end{equation}
where $\beta$ is the regularization weight and $r(y)$ is the reward function learned using
the Bradley-Terry model on the preference dataset of generating $y$.

\paragraph{Direct Preference Optimization (DPO)} DPO eliminates the training of reward model. 
It derives a loss on the current policy $\pi_{\theta}$ ($y_w$, $y_l$ omitted):
\begin{equation}
\label{eqn:dpo_loss}
\small
\mathcal{L}_{dpo} = - \log \sigma \left( \beta \log \frac{\pi_{\theta}(y_w)}{\pi_{\text{ref}}(y_w)} - \beta \log \frac{\pi_{\theta}(y_l)}{\pi_{\text{ref}}(y_l)} \right),
\end{equation}
i.e., the binary cross entropy with 
\begin{equation}
\small
    \hat{p}_{\theta}(y_w \succ y_l) = \sigma \left( \beta \log \frac{\pi_{\theta}(y_w)}{\pi_{\text{ref}}(y_w)} - \beta \log \frac{\pi_{\theta}(y_l)}{\pi_{\text{ref}}(y_l)} \right),
\end{equation}
and target $p(y_w \succ y_l) = 1$. We describe more PO methods in details in Appendix~\ref{appendix:po}.

\section{Methodology}

\subsection{Problem Formulation} 
\label{sec:promblem}
Formally, consider a query \(q\) and a corpus of text documents \(\mathcal{D}\). The goal is to produce an output \(\mathcal{S}\), where \(\mathcal{S}\) is a collection of \(n\) distinct statements: \(s_1, s_2, \ldots, s_n\). Each statement \(s_i\) is associated with a set of citations \(\mathcal{C}_i\). This set $\mathcal{C}_i$ is defined as \(\mathcal{C}_i = \{c_{i,1}, c_{i,2}, \ldots\}\), where each \(c_{i,j}\) is a document from the corpus \(\mathcal{D}\). For application purposes, the output from LLMs can be divided into individual statements using sentence boundaries. This approach is utilized because a single sentence typically encapsulates a coherent statement while maintaining brevity, facilitating easy verification. Regarding the citation format, citations are typically presented in square brackets, e.g., \textit{The sun is formed approximately 4.6 billion years ago}~\texttt{[1][2]}. However, it should be noted that these citations can be attributed to specific phrases as well, not just at the end of sentences.

Moreover, in this paper, we define \textit{generation hallucination} refers to a situation where the model generates content that is not based on factual information and  \textit{attribution hallucination} means that the statement corresponding to one citation is unfaithful or not supported by the referred source content.

\subsection{Overall Framework}
As shown in Figure~\ref{fig:framework}, we introduce the APO framework to apply preference learning for attribution task. The APO framework consists of the post-training procedure to ground the base model for attribution (\S\ref{sec:post-training}), and the preference optimization procedure to address both generation hallucination and attribution hallucination (\S\ref{sec:preference}). 

\subsection{Post-training}
\label{sec:post-training}
\begin{table}[]
\small
\centering
\begin{tabular}{@{}lccc@{}}
\toprule
\textbf{Dataset}  & \textbf{Source} & \textbf{\# Examples}\\ 
\midrule
\textbf{\textit{Post-training}} \\
EVIGSE      & Internet  & 3508    \\
ExpertQA     & Internet  & 906    \\
HAGRID     & Wiki  & 1301+615(dev)  \\
\midrule
\textbf{\textit{Preference Optimization}} \\
stanford\_alpaca     & Wiki  & 7741    \\
oasst1    & Wiki   & 2478    \\
asqa    & Wiki   & 2333    \\
sharegpt     & Wiki  & 2490    \\
wow    & Wiki   & 3689    \\
gpt4\_alpaca     & Wiki  & 6679    \\
flan\_v2     & Wiki  & 1693   \\
\midrule
\textbf{\textit{Test}} \\
ASQA    & Wiki   & 948    \\
StrategyQA     & Wiki  & 490    \\
ELI5    & Sphere   & 1000    \\
\bottomrule
\end{tabular}
\caption{Statistics of data collections used at different stages in the APO framework.}
\vspace{-4mm}
\label{tab:dataset}
\end{table}

The goal of post-training procedure is to ensure that given a specific question \(q\) and a corpus of text documents \(\mathcal{D}\), the model can be successfully instructed to generate answer \(\mathcal{S}\) and add citation \(\mathcal{C}_i\) for each statement $s_i$ in its response when necessary. 
\paragraph{Data Collection} We construct the post-training data from training sets using existing attribution datasets including EVIGSE~\cite{DBLP:conf/emnlp/LiuZL23}, ExpertQA~\cite{DBLP:journals/corr/abs-2309-07852} and HARGID~\cite{DBLP:journals/corr/abs-2307-16883}. We select these datasets because they are high-quality attribution datasets with diverse domains and sources annotated by human experts or powerful LLMs. After preprocessing and formatting, the final post-training data collection includes 6,330 samples. The pre-processing details are shown in Appendices~\ref{appendix:Datasets-Statistics} and~\ref{appendix:Post-training-Templates}, and the statistics of training data are shown in Table~\ref{tab:dataset}.

\paragraph{Training} After that, instruction \(\mathcal{I}_{post}\), documents \(\mathcal{D}\) and question  \(q\) are formatted to be the input while answer \(\mathcal{S}\) composed of multiple statements is formatted as output. We tune the model using autoregressive language modeling objectives, resulting in initial generator $M_g$.

\subsection{Preference Optimization}
\label{sec:preference}
\begin{algorithm}[t]
\small
\caption{Preference data sampling and labeling}
\begin{algorithmic}[1]

\State \textbf{Input} Queries $Q$, Critic $M_c$, Generator $M_g$, Retriever $R$
\State \textbf{Output} Output initialized preference dataset $P_{init}$

\State $P_{init}$ = \{\}
\For{$q\in Q$}
    \State Retrieve top-$k$ passages $D$ using $R$ given $q$
    \State Predicts \textit{relevant label} $\mathcal{L}_{rel} \in \{0, 1\}$ using critic $M_c$ for each $d$ in $D$ given $q$, $d$
    \State Generate \(\mathcal{S}\) constructed by statements $\{s_1,s_2,..,s_n\}$, using generator $M_g$ given $(\mathcal{I}_{post}, q,D)$.
    
    \For{$s_i\in \mathcal{S}  $}
        \State Predicts \textit{supported label} $\mathcal{L}_{sup} \in \{0, 1\}$ for each $c_{i,j}$ using critic $M_c$ given $q$, $d \leftrightarrow c_{i,j}$, $s_i$
    \EndFor
    \State Add augmented $(q, D, \mathcal{S}, \mathcal{C}, \mathcal{L}_{rel}, \mathcal{L}_{sup})$ to $P_{init}$
\EndFor
\end{algorithmic}
\label{preference_data_collection_step1}
\end{algorithm}

In this section, we describe our preference optimization procedure to enable a model-agnostic approach for improving the quality of generated responses. First, considering the cost of labeling preference data, we devise an automatic data collection algorithm motivated by errors where previous models may have misattributed. Second, we propose a progressive preference optimization approach to amplify the preference signal by using synthesized preference pairs. We further apply the experience replay to alleviate the over-fitting and text degradation phenomenon due to the distribution shift introduced by automatic data generation.

\paragraph{Automatic Data Collection}

In general, attributed text generation should be both relevant and supported~\cite{DBLP:journals/corr/abs-2310-11511}. Being relevant needs the reference document in the answer to be helpful in handling the question. It is used to measure whether \(\mathcal{C}\)  provides useful information to solve \(q\). Being supported asks the generated text be grounded on the reference documents. It is used to measure whether all of the verification-worthy statements in \(\mathcal{S}\) are supported by \(\mathcal{C}\).

Following the requirements above, we first get initial responses and related labels for each query with the Algorithm~\ref{preference_data_collection_step1}.  
The query comes from multiple open domain tasks or high-quality instruction data sets shown in Table~\ref{tab:dataset}. The source of retrieved documents is English Wikipedia. The retriever $R$ we use here is \texttt{gtr-t5-large}\footnote{\url{huggingface.co/sentence-transformers/gtr-t5-large}}.
The objective is to generate the attributed text with \textit{relevant} and \textit{supported} labels for related documents using the critic model $M_c$. Here, we use pre-trained \texttt{selfrag\_llama2\_7b}\footnote{\url{huggingface.co/selfrag/selfrag_llama2_7b}} as $M_c$ in~\citet{DBLP:journals/corr/abs-2310-11511} because it can give fine-grained feedback using reflection tokens.

After that, we generate preference pairs using an automatic collection algorithm. 
Specifically, we determine whether the citations $\mathcal{C}_i$ of each statement $s_i$ of query $q$ are all related to it based on the relevant tags. If it is all relevant, we add the current statement and its preceding statements $\mathcal{S}_i$ to the set $P_{tmp}$ for subsequent processing. For example, if $s_2$ meets the requirement, we add $\{s_1, s_2\}$ to $P_{tmp}$. The motivation here is that we want to select the statements that can answer the question based on the document as the initial set.
Then, for each entry in $P_{tmp}$, we first retrieve another top-$m$ ($m \gg k$) documents and filter them into 10 irrelevant documents $D_{ir}$  scored by relevant logits predicted by $M_c$. If all documents in $D_{ir}$ are relevant, we use the last 10 documents as $D_{ir}$.
After that, we generate the positive and negative pair for each statement $s_i \in P_{tmp}$. There are two situations: the statement $s_i$ is fully supported by $\mathcal{C}_i$ and otherwise. For the first situation, we first expand $\mathcal{C}_i$ with supported document by second judgment in $D$ using $M_c$. Then, we generate one positive statement using $q$, $\mathcal{S}_{i-1}$ and new $\mathcal{C}_i$ and two negative statements using $q$, $\mathcal{S}_{i-1}$, $D_{ir}$ and $q$, $\mathcal{S}_{i-1}$, new $\mathcal{C}_i$, error instruction $e$ respectively. 
Thus, there are two preference pairs in this context.
For the second situation, we generate one positive statement using $q$, $\mathcal{S}_{i-1}$ and new $\mathcal{C}_i$ and one negative statement using  $q$, $\mathcal{S}_{i-1}$,  $D - \mathcal{C}_i$, error instruction $e$. The full procedure is shown in Algorithm~\ref{preference_data_collection_step2}.

In the generation of negative samples, we use the error instruction $e \in \mathcal{E}$, which defines two types: \textit{irrelevant but supported} means the generated text \(s_i\) is grounded on unhelpful reference documents \(\mathcal{C}_i\), while \textit{relevant but unsupported} further has three fine-grained subtypes: 1) \textit{fabricated statement} refers to the generated text contains facts or information that cannot be derived from reference documents; 2) \textit{mistaken synthesis} means that several reference documents are used, but facts or logics are mistakenly intermingled. The generated text thus contains factual error or logic error; 3) \textit{unintentional omission} means that reference documents are used, but the key points are incomplete. There are no factual errors in generated text, but some information is omitted. 
The \textit{irrelevant but supported} error derives from attribution hallucination, whereas the \textit{relevant but unsupported} error is the result of generation hallucination. Note that irrelevant and unsupported errors are not included, since it is more like easy negatives. The details of error instructions are in Appendix~\ref{appendix:details_instructions}.

\paragraph{Progressive Preference Optimization}
To reinforce the preference feature and alleviate sparse reward problem~\cite{DBLP:journals/corr/abs-2307-04964,DBLP:journals/corr/abs-2305-20050}, we propose a progressive preference optimization method. Considering generations can be separated into several consecutive statements, each statement may contain hallucinations at all. The entire response-level reward preference modeling performs in the global context and potentially oversights the fine-grained deterministic preferences we constructed. Hence, we use fine-grained statement-level reward to perform preference optimization to update the model in a more effective and efficient way.
Formally, assuming that deterministic preference is performed at statement-level, we can rewrite the preference optimization loss in  Eqn. (\ref{eqn:dpo_loss}) as follows ($-\log\sigma$ omitted):
\begin{equation}
\small
\begin{aligned}
    \mathcal{L} &\triangleq \beta\log\frac{\pi_\theta(y_w)}{\pi_{ref}(y_w)}-\beta\log\frac{\pi_\theta(y_l)}{\pi_\theta(y_w)} \\
    &=\beta\log\frac{\sum_i\pi_\theta(s_i^w|s_{:i-1}^w)}{\sum_i\pi_{ref}(s_i^w|s_{:i-1}^w)}-\beta\log\frac{\sum_j\pi_\theta(s_j^l|s_{:j-1}^l)}{\sum_j\pi_{ref}(s_j^l|s_{:j-1}^l)} \\
    &=\beta\sum_i\log\frac{\pi_\theta(s_i^w|s_{:i-1}^w)}{\pi_{ref}(s_i^w|s_{:i-1}^w)}-\beta\sum_j\log\frac{\pi_\theta(s_j^l|s_{:j-1}^l)}{\pi_{ref}(s_j^l|s_{:j-1}^l)} \\
    &=\beta\sum_i\left(\log\frac{\pi_\theta(s_i^w|s_{:i-1}^w)}{\pi_{ref}(s_i^w|s_{:i-1}^w)}-\log\frac{\pi_\theta(s_i^l|s_{:i-1}^l)}{\pi_{ref}(s_i^l|s_{:i-1}^l)}\right).
\end{aligned}
\end{equation}
The progressive preference optimization loss can be further written as follows ($-\log\sigma$ omitted):


\begin{equation}
\small
\begin{aligned}
    \mathcal{L} &\triangleq \mathop{\mathbb{E}}_{(s_i^w,s_i^l\sim D)}\left(\beta\log\frac{\pi_\theta(s_i^w)}{\pi_{ref}(s_i^w)}- 
    \beta\log\frac{\pi_\theta(s_i^l)}{\pi_{ref}(s_i^l)}\right) \\
    &=\mathop{\mathbb{E}}_{(y_w,y_l\sim D)}\frac{1}{n}\sum_i\left(\beta\log\frac{\pi_\theta(s_i^w)}{\pi_{ref}(s_i^w)}- \beta\log\frac{\pi_\theta(s_i^l)}{\pi_{ref}(s_i^l)}\right).
\end{aligned}
\end{equation}
The main difference between vanilla preference optimization in Eqn. (\ref{eqn:dpo_loss}) and progressive preference optimization is that the latter contains an implicit mean pooling procedure when implementing the preference optimization loss.

Furthermore, the directed preference optimization may face the challenges of overfitting to some deterministic preference due to weak KL constraint~\cite{DBLP:journals/corr/abs-2310-12036}. Hence, we propose to leverage experience replay~\cite{DBLP:conf/nips/RolnickASLW19} as learning with rehearsal to alleviate the over-fitting phenomenon. The idea of replaying experience typically stores a few old training samples within a small memory buffer. Therefore, we iteratively add post-training autoregressive language modeling loss to the preference optimization procedure in a fixed interval, resulting in final generator $M_p$.

\subsection{Inference and Refinement}
During inference, for query $q$, $D$ is first retrieved and then sent to $M_p$ output to the final answer $\mathcal{S}_{init}$ consists of $n$ statements.
As there may not be all statements correctly attributing documents, we additionally perform the post-hoc refinement after the original generation. We maintain a collection of citations $\mathcal{C}_{tmp}$.
Starting from the last statement of $\mathcal{S}_{init}$, if the current $s_i$ has the citations, update the $\mathcal{C}_{tmp}$ to the citations of the current $s_i$; if the current $s_i$ does not have a citation, add the current citation set $\mathcal{C}_{tmp}$ to this statement until all $n$ statements have been traversed. Then we concatenate these $n$ statements together as the final answer $\mathcal{S}$.

\section{Setup}
\begin{table*}[t]
\centering
\setlength\tabcolsep{2pt}
\footnotesize
\begin{tabular}{@{}lcccccccccccccccc@{}}
\toprule
\multicolumn{1}{c}{\multirow{3}{*}{\textbf{Dataset \& Metrics}}}                                                                    & \multicolumn{4}{c}{\textbf{ASQA}}                               & \multicolumn{4}{c}{\textbf{StrategyQA}}                                                        & \multicolumn{4}{c}{\textbf{ELI5}}              \\ \cmidrule(lr){2-14}
\multirow{2}{*}{} & \textbf{Correct} & \multicolumn{3}{c}{\textbf{Citation}}        & \textbf{Correct}         & \multicolumn{3}{c}{\textbf{Citation}}        & \textbf{Correct} & \multicolumn{3}{c}{\textbf{Citation}}  \\ \cmidrule(lr){2-14}
    & \textbf{EM-R}  & \textbf{Rec} & \textbf{Prec} & \textbf{F1}   & \textbf{ACC}   & \textbf{Rec} & \textbf{Prec} & \textbf{F1}   & \textbf{Claim}   & \textbf{Rec} & \textbf{Prec} & \textbf{F1}   &                                   &                                                                                \\ 
\midrule
\textsc{ICLCite}~\cite{DBLP:conf/emnlp/GaoYYC23} & 35.2  & 38.4  & 39.4 & 38.9  & \textbf{65.5} & 20.6 & 33.1 & 25.4 & 13.4 & 17.3 & 15.8  & 16.5    \\
\textsc{PostCite}~\cite{DBLP:conf/emnlp/GaoYYC23}   & 25.0  & 23.6  & 23.6 & 23.6 & 64.3 & 8.7 & 8.7 & 8.7 & 7.1 & 5.7 & 5.8 & 5.8 \\ 
\textsc{PostAttr}~\cite{ye2023effective}     & 25.0  & 33.6  & 33.6 & 33.6  & 64.3 & 12.5 & 12.5 & 12.5 & 7.1 & 12.2 & 12.2 & 12.2  \\ 
Self-RAG~\cite{DBLP:journals/corr/abs-2310-11511}     & 31.7  & 70.3  & \textbf{71.3}  & 70.8 & 62.1 & 31.4 & 36.5 & 33.8 & 10.7 & 20.8 & 22.5 & 21.6 \\ 
AGREE~\cite{ye2023effective}     & 39.4  & 64.0  & 66.8  & 65.4 & 64.6 & 30.2 & 37.2 & 33.3 & 9.4 & 21.6 & 16.0 &   18.4\\ 
\midrule
APO (only post-training) \hspace{-10pt}    & 36.6 & 65.0 & 62.1 & 63.5 & 62.5  & 30.7  & 30.1 & 30.4 & 13.0 & 18.5 & 17.9 & 18.2 \\
APO (our method) \hspace{-10pt}     & \textbf{40.5} &\textbf{72.8} & 69.6 & \textbf{71.2} & 61.8  & \textbf{40.0}  & \textbf{39.1} & \textbf{39.6}& \textbf{13.5} & \textbf{26.0} & \textbf{24.5} & \textbf{25.2} \\ 
\bottomrule
\end{tabular}

\caption{The performance comparison between our method and extensive baselines. Experiments are evaluated on ASQA~\cite{DBLP:conf/emnlp/StelmakhLDC22}, StrategyQA~\cite{DBLP:journals/tacl/GevaKSKRB21} and ELI5 dataset~\cite{DBLP:conf/acl/FanJPGWA19}. For most baselines, we use the results of previous works~\cite{DBLP:conf/emnlp/GaoYYC23,ye2023effective}. 
}
\vspace{-4mm}
\label{tab:main_result}
\end{table*}
\subsection{Datasets and Evaluation Metrics}
\paragraph{Dataset} We mainly focus on attributable long-form question-answering (QA) task using ASQA dataset and ELI5 subsets from~\citet{DBLP:conf/emnlp/GaoYYC23}. In addition to these factoid long-form QA tasks, we test the generation quality on StrategyQA dataset~\cite{DBLP:journals/tacl/GevaKSKRB21} which focuses on open-domain QA where the required reasoning steps are implicit in the question. We use the official test set as our evaluation set.  

\paragraph{Metrics} Following~\citet{DBLP:conf/emnlp/GaoYYC23}, we report citation \textbf{recall}, \textbf{precision}, and \textbf{F1} which uses \textsf{TRUE}~\cite{DBLP:conf/naacl/HonovichAHTKCSS22} as the attribution evaluation model $\phi$ to automatically examine whether the cited documents entail the model generation. 
For ASQA dataset, we report the \textbf{recall of correct short answers} (EM-R) by checking whether the short answers (provided by the dataset) are exact substrings of the generation. 
For ELI5 dataset, we report the \textbf{claim recall} (Claim) to check whether the model output entails the sub-claims, that are generated by \texttt{text-davinci-003}~\cite{DBLP:conf/nips/Ouyang0JAWMZASR22}. 
For StrategyQA dataset,  we report the \textbf{accuracy} for task performance.

\subsection{Competitive Methods}
We compare APO with several baselines. For each baseline, we use \texttt{gtr-t5-large} as our retriever.

\textbf{In-Context Learning} (\textsc{ICLCite}): We prompt LLMs with few-shot examples, each consisting of a query, a set of retrieved documents and an answer with inline citations. The LLMs can in-context learn from the examples and generate grounded responses for the test query and retrieved documents.
 
\textbf{Post-Hoc Cite} (\textsc{PostCite}): Given query $q$, we first instruct LLMs to answer $q$ \textit{without} retrieved documents. Then, we use the attribution evaluation model $\phi$ to link each statement to the most relevant document retrieved by the query.

\textbf{Post-Hoc Attribute} (\textsc{PostAttr}): Instead of citing the most relevant document, for each statement, we further retrieve a set of \textit{k} documents and then use the $\phi$ to link to the document that maximally supports the statement by threshold.

\textbf{Self-RAG}~\cite{DBLP:journals/corr/abs-2310-11511}: Self-RAG is the state-of-the-art (SoTA) method that adaptively retrieves documents on-demand. It generates with reflection on retrieved documents and its generations by special token control.

\textbf{AGREE}~\cite{ye2023effective}: AGREE leverages test-time adaptation to reinforce unverified statements which iteratively improves the responses of LLMs.  It tunes a pre-trained LLM to self-ground its response in retrieved documents using automatically collected data.

\subsection{Implementation Details}
If not specified, we retrieve the top 5 documents as the related documents to $q$ and we set the decoding temperature to 0.01 during inference. For the post-training, we tune the model for 2 epochs with a learning rate of 5e-5. For the preference optimization, we tune the model with LoRA~\cite{DBLP:conf/iclr/HuSWALWWC22} for 1 epoch, and we set alpha to 2 and lora ranks to 16. We set $m$ to 100. We use \texttt{llama-2-13b-base}~\cite{DBLP:journals/corr/abs-2307-09288}  for fair comparison. We run all the experiments on NVIDIA A100 80G GPUs.

\section{Results}

\subsection{Main Result}
Table~\ref{tab:main_result} shows the comparison results of APO with other baselines on three datasets.  In terms of correctness and citation quality, our method outperforms the baselines on all three datasets. It shows that APO has better overall generation performance in various scenarios. Specifically, our method outperforms Self-RAG by 8.8 points on the EM-R metric. We speculate that this inconsistency stems from the difference between coherent generation and step-wise generation in Self-RAG. Our method also shows consistent improvements over AGREE across multiple benchmarks which suggests that APO can more effectively exploit the power of LLM to enhance retrieval. APO can be used to complement these active or adaptive retrieval-based methods and we leave it for future work. Compared to the post-training baseline, the preference optimization shows further improvement with an 8.0 average increased citation F1. Furthermore, we observe a trade-off between correctness and citation quality in several baselines including Self-RAG and AGREE, possibly due to the generation hallucination and attribution hallucination defined in \S\ref{sec:promblem}. In contrast, APO helps to deal with these hallucinations and performs well in terms of both correctness and citation quality.

\subsection{Ablation Study}
\begin{table}[]
\small
\centering
\resizebox{\linewidth}{!}{ 
\begin{tabular}{@{}lcccc@{}}
\toprule
            \textbf{Method }          & \textbf{EM-R} & \textbf{Rec} & \textbf{Prec} & \textbf{F1} \\ 
\midrule
Our Method    & 36.6  & 65.0  & 62.1  & 63.5    \\
\hdashline
\ \  w/o asqa    & 38.8  & 71.7  & 67.2  & 69.4    \\
\ \  w/o hallucinated statement & 40.4 & 69.3 & 65.3 & 67.3 \\
\ \  w/o mistaken synthesis & 40.2 & \textbf{73.4} & 69.2 & 71.2 \\
\ \  w/o unintentional omission   & 39.1 & 72.7 & 68.2 & 70.4 \\
\hdashline
\ \  w/ response-level PO & 38.9 & 69.1 & 65.1 & 67.1 \\
\ \  w/ statement-level PO\hspace{-20pt}    & \textbf{40.5}  & 72.8  & \textbf{69.6}  & \textbf{71.2}    \\
\bottomrule
\end{tabular}}
\caption{Ablation study on the ASQA dataset. We ablate not only the source and predefined error type used to construct PO data, but also the training strategy.}
\vspace{-4mm}
\label{tab:data_comparsion}
\end{table}

We evaluate the effectiveness of each predefined error type and the results are shown in Table~\ref{tab:data_comparsion}. Specifically, we perform progressive PO on the model after post-training and remove data corresponding to a predefined type. We observe that without data corresponding to hallucinated statement error, citation F1 drops significantly which suggests that our approach improves the groundedness of the model. Mistaken synthesis error seems to contribute little to performance improvement, but we observe that it can help improve groundedness under human evaluation (\S\ref{sec:ea}). Without unintentional omission error, the model shows poor generation quality. This means that the model may generate incomplete answers. 

Moreover, we perform an ablation study on the training strategy of preference optimization. We find that the model can also be improved under the response-level preference optimization method such as vanilla DPO, but the improvement is slightly less. In addition, we ablation the PO by removing the ASQA questions from our preference data. Note that we construct the preference data based on the training set of ASQA, and use its test set for evaluation. We have verified and guaranteed that there is no data overlap between the two. We find that the generation quality and citation quality have decreased. We attribute it to high-quality in-domain questions in ASQA as a long-form question answering dataset.  

\subsection{Different Prompting Strategy}


\begin{table}[]
\footnotesize
\centering
\begin{tabular}{@{}lcccc@{}}
\toprule
   \multicolumn{1}{c}{\multirow{2}{*}{\textbf{Method \& Metrics}}}                      & \multicolumn{4}{c}{\textbf{ASQA}}                          \\ 
\cmidrule(l){2-5} 
                       & \textbf{Correct} & \textbf{Rec} & \textbf{Prec} & \textbf{F1} \\ 
\midrule
\textbf{\texttt{llama-2-13b-chat}} \\
\midrule
\textsc{Vanilla}(5-psg) & 32.6 & 60.0 & 52.1 & 55.8\\
\textsc{Summ}(10-psg) & 42.9 & 58.7 & 50.4 & 54.2 \\
\textsc{Snippet}(10-psg) & 41.3 & 57.4 & 52.1 & 54.6 \\
\textsc{Oracle}(5-psg) & 41.4 & 54.5 & 52.9 & 53.7\\
\midrule
\textbf{Our method} \\
\midrule
\textsc{Vanilla}(5-psg) & 40.5  & \textbf{72.8}  & \textbf{69.6}  & \textbf{71.2 }\\
\textsc{Summ}(10-psg) & 42.7  & 60.9  & 53.4  & 56.9 \\
\textsc{Snippet}(10-psg) & 42.3  & 57.8  & 51.6 & 54.5   \\
\textsc{Oracle}(5-psg) & \textbf{52.4}  & 70.5  & 66.2  & 68.3 \\ 
\midrule
     \multicolumn{1}{c}{\multirow{2}{*}{\textbf{Method \& Metrics}}}                   & \multicolumn{4}{c}{\textbf{ELI5}}                       \\ 
\cmidrule(l){2-5} 
                       & \textbf{Correct} & \textbf{Rec} & \textbf{Prec} & \textbf{F1} \\ 
\midrule
\textbf{\texttt{llama-2-13b-chat}}  \\
\midrule
\textsc{Vanilla}(5-psg) & 12.1 & 16.4 & 19.7 & 17.9 \\
\textsc{Summ}(10-psg) & 6.1 & 9.9 & 14.3 & 11.7 \\
\textsc{Snippet}(10-psg) & 11.9 & 29.4 & 28.6 & 29.0 \\
\textsc{Oracle}(5-psg) & 16.9 & 21.4 & 27.3 & 24.0 \\
\midrule
\textbf{Our method} \\
\midrule
\textsc{Vanilla}(5-psg) & 13.5  & 26.0  & 24.5  & 25.2 \\
\textsc{Summ}(10-psg) & 12.7  & \textbf{37.8}  & \textbf{35.7}  & \textbf{36.7} \\
\textsc{Snippet}(10-psg) & 14.2  & 37.6  & 34.8  & 36.1   \\
\textsc{Oracle}(5-psg) & \textbf{21.7}  & 32.6  & 30.8  & 31.7 \\ 
\bottomrule
\end{tabular}
\caption{Comparisons with different retrieval context.}
\vspace{-4mm}
\label{tab:doc_quality}
\end{table}

We explore applying APO to four prompting strategies~\cite{DBLP:conf/emnlp/GaoYYC23}: 1) \textsc{Vanilla} that provides the top-5 retrieved documents for each question. It is our default setting. 2) \textsc{Summ} that provides summaries instead of the full text of the top-10 retrieved documents for each question. 3)  \textsc{Snippet} that provides snippets instead of the full text of the top 10 retrieved documents for each question. 4) \textsc{Oracle} that provides 5 gold documents for each question. We use \texttt{llama-2-13b-chat} as the comparison method because it has impressive instruction following ability and moderate size. As shown in Table~\ref{tab:doc_quality}, we find that in most cases, APO achieves better performance than baseline. For example, APO under \textsc{Vanilla} and \textsc{Oracle} settings performs best in Citation F1 on ASQA, while it under \textsc{Summ} and \textsc{Snippet} settings in ELI5 has improved Citation F1. It shows that the format of the context has an impact on attribution task.


\subsection{Different PO Methods}
\begin{table}[t]
\small
\centering
\resizebox{\linewidth}{!}{ 
\begin{tabular}{@{}lcccc@{}}
\toprule
        \multicolumn{1}{c}{\multirow{2}{*}{\textbf{Method \& Metrics}}}               & \multicolumn{4}{c}{\textbf{ASQA}}                          \\ 
\cmidrule(l){2-5} 
                       & \textbf{Correct} & \textbf{Rec} & \textbf{Prec} & \textbf{F1} \\ 
\midrule
APO (only post-training)    &  36.6  &  65.0  & 62.1  & 63.5    \\
  \hdashline
\ \  w/ Positive statement SFT    & 29.0  & 66.7  & 56.8  & 61.4    \\
\ \  w/ IPO~\cite{DBLP:journals/corr/abs-2310-12036}    & 39.9  & 72.7  & 69.2  & 70.9   \\
\ \  w/ SLiC~\cite{SLiC-HF}   & 40.1  & 72.5  & 69.1  & 70.8 \\
\ \  w/ KTO~\cite{ethayarajh2023halos}    & 39.8  & 72.5  & 68.7  & 70.5    \\ 
\ \  w/ Progressive PO\hspace{-20pt}    & \textbf{40.5}  & \textbf{72.8}  & \textbf{69.6}  & \textbf{71.2}    \\
\midrule
         \multicolumn{1}{c}{\multirow{2}{*}{\textbf{Method \& Metrics}}}              & \multicolumn{4}{c}{\textbf{ELI5}}                       \\ 
\cmidrule(l){2-5} 
                       & \textbf{Correct} & \textbf{Rec} & \textbf{Prec} & \textbf{F1} \\ 
\midrule
APO (only post-training)    & 13.0  & 18.5  & 17.9  & 18.2     \\
\hdashline
\ \  w/ Positive statement SFT    & 10.6  & \textbf{34.5}  & \textbf{30.8}  & \textbf{32.5}    \\
\ \  w/ IPO~\cite{DBLP:journals/corr/abs-2310-12036}    & 13.5  & 26.5  & 24.8  & 25.6   \\
\ \  w/ SLiC~\cite{SLiC-HF}   & 13.7 & 30.7 & 22.0 & 25.6 \\
\ \  w/ KTO~\cite{ethayarajh2023halos}    & \textbf{14.3}  & 24.7  & 26.5  & 25.6    \\
\ \  w/ Progressive PO\hspace{-20pt}   & 13.5  & 26.0  & 24.5  & 25.2    \\ 
\bottomrule
\end{tabular}
}
\caption{Comparisons with different preference method.}
\label{tab:preference_comparsion}
\end{table}

Table~\ref{tab:preference_comparsion} illustrates the results of different direct preference optimization methods adopted by $M_p$. We include a SFT baseline to tune the $M_g$ using the positive part in the chosen preference pairs that we created. We observe that our method can be transferred to several different preference optimization methods, but the performance swings in several metrics. All preference optimization methods have performance boosts compared with the post-training baseline and the SFT baseline. It shows that preference optimization can help improve the generation quality to some extent. 

\subsection{Error Analysis}
\label{sec:ea}
\begin{table}[]
\centering
\footnotesize
\begin{tabular}{@{}lc@{}}
\toprule
\textbf{Error Type}  & \textbf{\# Proportion (\%)}\\ 
\midrule
Attribution hallucination & 26.4 \\
\hdashline
Generation hallucination & \\
\ \ - Fabrication  & 48.4    \\
\ \ - Omission  & 18.7    \\
\ \ - Synthesis  & 6.5  \\
\bottomrule
\end{tabular}
\caption{Error types of the proposed methods.}
\vspace{-4mm}
\label{tab:error_proportion}
\end{table}

We conduct human evaluation of model response on ASQA dataset. 
Specifically, we collect 50 samples that contain errors judged by the attribution evaluation model $\phi$. We then perform a detailed manual review of these samples to identify error types. Our evaluation results are shown in Table~\ref{tab:error_proportion}.
We find that nearly half of the errors are of fabrication error. We reveal that the model either generated text not supported by the reference documents or incorrectly attributed information to irrelevant documents. In certain instances, hallucinations are due to the documents with low quality. For example, some documents are truncated, and the model attempts to complete or extrapolate the incomplete text. Additionally, we notice omission errors on both generated text and citation where the model fails to generate necessary citations to substantiate its statements. Although synthesis errors are less common,  we observe some cases which model conflated information from multiple documents and generated counterfactual statements.
The case study is shown in Appendix~\ref{appendix:ea}.


\section{Conclusion}
This paper introduces the APO framework for attributed text generation. We treat attribution as a preference learning task, utilizing curated post-training collections and an automated synthesis algorithm to reduce manual labeling costs. Experiments on three datasets demonstrate the effectiveness of APO which achieves leading citation F1 and improved response quality.  
Future work can explore extending APO to real-world applications.
\clearpage
\section*{Limitation}
We aim to improve the credibility and reliability of content generated by LLMs using the APO framework. However, it faces limitations such as the narrow scope of datasets used, which may not fully represent the diversity of real-world applications~\cite{DBLP:conf/emnlp/LiuZL23}. The generalization capabilities of the model are also a concern, as the automatic generated data may not cover all scenarios of hallucination. While addressing the high cost of data labeling, the scalability and economic feasibility in larger datasets remain unexplored. The approximation of human citation processes may not capture all the complexities of scholarly writing, and its reliance on external sources raises concerns about the quality and availability of these sources. Potential biases in training data and synthesized data could lead to biased outputs~\cite{DBLP:conf/acl/WangKMLSKH23,DBLP:conf/emnlp/HuCZ15}. The robustness of the framework against deliberate hallucination and its adaptability to rapidly evolving NLP fields are not fully assessed, highlighting areas for future improvement and research in enhancing LLM reliability.
\section*{Ethical Statement}
The ethical considerations surrounding the use of LLMs that generate citations encompass a range of concerns, including the risk of increased trust without verification, challenges in time-critical decision-making, the assumption of inherent trustworthiness, and copyright issues. Ethically, it is crucial to encourage users to critically engage with and verify machine-generated content to mitigate misinformation and hallucinations. Additionally, recognizing the limitations and potential legal challenges related to copyright when using such attributions is essential. Addressing these ethical issues and educating users on the potential pitfalls becomes increasingly important to ensure the responsible and informed use of text generated by LLMs.

\bibliography{custom_pl}

\clearpage
\appendix

\section{Details about Preference Optimization Methods}
\label{appendix:po}
\paragraph{Reinforcement Learning form Human Feedback (RLHF)}~\cite{DBLP:conf/nips/Ouyang0JAWMZASR22} uses reward-model-based reinforcement learning algorithm to learn the optimal policy. It first learns a reward model from the preference data, then uses an on-policy PPO algorithm~\cite{DBLP:journals/corr/SchulmanWDRK17} to maximize the learned reward. The reward is learned to use Bradley-Terry model~\cite{bradley1952rank}, which assumes the preference score can be approximated by substituted with point-wise reward. This assumption may lead to an approximation error when preference is deterministic. The PPO algorithm is used on data sampled from generating policy, which may have a different support or distribution drift from preference data, the learned reward model inference on the out-of-distribution data may reduce the accuracy. The process of RLHF needs to train reward model and on-policy PPO algorithm which is complex, time-consuming, and unstable. 

\paragraph{Direct Preference Optimization (DPO)}~\cite{DBLP:journals/corr/abs-2305-18290} combines off-policy algorithm and Bradley-Terry model to directly learn the generating policy from preference data. The off-policy algorithm is based on KL-regularization reward maximization from off-RL community, which is data efficient, stable and eliminating the need for a reward model. When preference is deterministic which occurs in most cases, the reward of Bradley-Terry model is undefined, which leads to ignoring the KL-regularization term and overfitting the preference dataset. 

\paragraph{Identity-mapping Preference Optimization (IPO)}~\cite{DBLP:journals/corr/abs-2310-12036} claims when preferences are deterministic or near deterministic, DPO will lead over-fitting to the preference dataset at the expense of ignoring the KL-regularation term. 
To optimize the objective, IPO derives an off-policy loss on empirical dataset:
\begin{equation}
h_{y_w,y_l}^{\pi_{\theta}} = \log \frac{\pi_{\theta}(y_w)}{\pi_{\text{ref}}(y_w)} - \log \frac{\pi_{\theta}(y_l)}{\pi_{\text{ref}}(y_l)},
\end{equation}
\begin{equation}
L_{\text{IPO}}(\theta, y_w, y_l) = \left( h_{y_w,y_l}^{\pi} - \frac{1}{2\beta} \right)^2.
\end{equation}
That means IPO loss will always regularize $\pi_\theta$ towards
$\pi_{\text{ref}}$ by controlling the gap between the log-likelihood ratios $\log\frac{\pi_\theta(y_w|x)}{\pi_\theta(y_l|x)}$
and $\log\frac{\pi_{\text{ref}}(y_w|x)}{\pi_{\text{ref}}(y_l|x)}$.

\paragraph{Kahneman-Tversky Optimization (KTO)}~\cite{ethayarajh2023halos}  directly maximizes the utility of LLM generations instead of maximizing the log-likelihood of preferences by introducing a Kahneman-Tversky Optimization loss. KTO does not need preference pairs and only knowledge of whether output is desirable or undesirable for a given input.

\paragraph{Sequence Likelihood Calibration (SLiC)}~\cite{SLiC-HF} uses calibrated likelihood of model-generated sequences to better align with reference sequences in the model’s latent space. It tries to alleviate the problem of MLE that gives probability mass to sparsely observed target sequences, which is used to calculate reward in DPO.

\section{Details about Pre-processing}
\label{appendix:Datasets-Statistics}
\begin{table*}[]
\small
\centering
\resizebox{\linewidth}{!}{ 
\begin{tabular}{@{}lccccc@{}}
\toprule
            \textbf{Dataset}          & \textbf{\# Sample} & \textbf{Avg. Query Length} & \textbf{Avg. Response Length} & \textbf{Avg. Statements} & \textbf{Avg. Citations} \\ 
\midrule
EVIGSE    & 3508  & 51.03  & 379.05  & 4.32 & 3.19    \\
ExpertQA & 906 & 106.90 & 999.84 & 7.16 & 5.67  \\
HARGID(train) & 1301 & 38.55 & 368.22 & 4.62 & 2.85 \\
HARGID(dev) & 615 & 40.43 & 292.46 & 3.63 & 2.54 \\
\bottomrule
\end{tabular}}
\caption{Details for our post-training data after pre-processing.}
\label{tab:pt_data}
\vspace{-4mm}
\end{table*}

For ExpertQA dataset, we remove samples whose 1) citations attribute to empty references; 2) documents contain different document IDs but same context. For EVIGSE dataset, we remove samples whose 1) citation attribute to ``None'' references; 2) do not have reference documents. We further normalize the ``supported'' label and the citation format for these datasets. The details of each dataset we used for post-training procedure after pre-processing are shown in Table~\ref{tab:pt_data}.

\section{Post-training Templates}
\label{appendix:Post-training-Templates}
\begin{table*}[h!]
\begin{tcolorbox}
{\bf Input}\\ 
Write an accurate, engaging, and concise answer for the given question using only the provided documents (some of which might be irrelevant) and cite them properly. Use an unbiased and journalistic tone. Always cite for any factual claim. When citing several search results, use [1][2][3]. Cite at least one document and at most three documents in each sentence. If multiple documents support the sentence, only cite a minimum sufficient subset of the documents.

Question: \{\{question\}\}

Document [1](Title: \{\{title 1\}\}): \{\{context 1\}\}

Document [2](Title: \{\{title 2\}\}): \{\{context 2\}\}

Document [3](Title: \{\{title 3\}\}): \{\{context 3\}\}

...

Document [n](Title: \{\{title n\}\}): \{\{context n\}\}

Answer: 
\tcblower
{\bf Output}  \\
\{\{output\}\} \\
\end{tcolorbox}
\caption{Post-training Template with instruction \(\mathcal{I}_{post}\)}
\label{tab:template}
\end{table*}

\begin{table*}[h!]
\begin{tcolorbox}
{\bf Input}\\ 
Task: Your job is to write a high quality response with requirements as follows:\\
General: Given Request, incomplete response and evidence, continue write a single sentence as the next sentence of the unfinished response. If text in unfinished response is ``None'', you should start the response(the first sentence).\\
Detail: You should always use the facts from the evidences to propuse your response. Your response is correct and comprehensive, fully supported by the evidence we provided. **Don't use any evidence that can be directly retrieved from the evidences we provided**. No hallucinations, no factual errors, no logic errors.\\

Request: \{\{request\}\}

Evidence: 

Document [1](Title: \{\{title 1\}\}): \{\{context 1\}\}

Document [2](Title: \{\{title 2\}\}): \{\{context 2\}\}

Document [3](Title: \{\{title 3\}\}): \{\{context 3\}\}

...

Document [n](Title: \{\{title n\}\}): \{\{context n\}\}

Unfinished response: \{\{past statements\}\}

Next sentence(good):
\tcblower
{\bf Output}  \\
\{\{output\}\} \\
\end{tcolorbox}
\caption{Positive Template}
\label{tab:positive_template}
\end{table*}

\begin{table*}[h!]
\begin{tcolorbox}
{\bf Input}\\ 
Task: Your job is to write a low quality response with requirements as follows:\\
General: Given Request, incomplete response and evidence, continue write a single sentence as the next sentence of the unfinished response. If text in unfinished response is ``None'', you should start the response(the first sentence).\\
Detail: You will always ignore the evidence. On one hand, you won't follow the evidence we provided, your response should be irrelevant to the evidence we provided. On the other hand, your response should be relevant to the unfinished response.\\

Request: \{\{request\}\}

Evidence: 

Document [1](Title: \{\{title 1\}\}): \{\{context 1\}\}

Document [2](Title: \{\{title 2\}\}): \{\{context 2\}\}

Document [3](Title: \{\{title 3\}\}): \{\{context 3\}\}

...

Document [n](Title: \{\{title n\}\}): \{\{context n\}\}

Unfinished response: \{\{past statements\}\}

Raw sentence(good): \{\{positive statement\}\}

Worse sentence(bad, ignore the evidence):
\tcblower
{\bf Output}  \\
\{\{output\}\} \\
\end{tcolorbox}
\caption{Negative, fabrication template}
\label{tab:hallu_template}
\end{table*}

\begin{table*}[h!]
\begin{tcolorbox}
{\bf Input}\\ 
Task: Your job is to write a low quality response with requirements as follows:\\
General: Given Request, incomplete response and evidence, continue write a single sentence as the next sentence of the unfinished response. If text in unfinished response is ``None'', you should start the response(the first sentence).\\
Detail: You should first, identify the relationships and entities in evidence; second, continue writing the next sentence of the response span with regard to the evidence. In your response, the relationships and entities should be mistakenly intermingled(you are making negative samples, we need low-quality data).\\

Request: \{\{request\}\}

Evidence: 

Document [1](Title: \{\{title 1\}\}): \{\{context 1\}\}

Document [2](Title: \{\{title 2\}\}): \{\{context 2\}\}

Document [3](Title: \{\{title 3\}\}): \{\{context 3\}\}

...

Document [n](Title: \{\{title n\}\}): \{\{context n\}\}

Unfinished response: \{\{past statements\}\}

Raw sentence(good): \{\{positive statement\}\}

Worse sentence(bad, entities in evidences mistakenly intermingled):
\tcblower
{\bf Output}  \\
\{\{output\}\} \\
\end{tcolorbox}
\caption{Negative, synthesis template}
\label{tab:syn_template}
\end{table*}

\begin{table*}[h!]
\begin{tcolorbox}
{\bf Input}\\ 
Task: Your job is to write a low quality response with requirements as follows:

General: Given Request, unfinished response and next sentence, omit some important points from the next sentence(good) and convert it into a worse response. Your converted worse response should be consistent with the unfinished response.\\

Request: List the ingredients needed to make a peanut butter and jelly sandwich

Unfinished response: 

Raw sentence(good): To make a peanut butter and jelly sandwich, you will need peanut butter, jelly or jam of your choice, and bread.

Worse sentence(bad, omit the evidence): To make a peanut butter and jelly sandwich, you will need peanut butter and bread.\\

Request: What are the three features of a cloud-based Database-as-a-Service (DBaaS)?

Unfinished response: The three main features of a cloud-based DBaaS are scalability, cost efficiency, and backups. Scalability allows you to increase or decrease the resources used by the DBaaS with ease.

Raw sentence(good): Cost efficiency is another important feature of a cloud-based DBaaS, as it allows you to pay for only the resources you need and eliminates the need for upfront hardware investments.

Worse sentence(bad, omit the evidence): Cost efficiency is another important feature of a cloud-based DBaaS, as it allows you to pay for only the resources you need.\\

Request: \{\{request\}\}

Unfinished response: \{\{past statements\}\}

Raw sentence(good): \{\{positive statement\}\}

Worse sentence(bad, omit the evidence):
\tcblower
{\bf Output}  \\
\{\{output\}\} \\
\end{tcolorbox}
\caption{Negative, omission template}
\label{tab:omission_template}
\end{table*} 

The post-training template we used follows the question answering template used by ~\citet{DBLP:conf/emnlp/GaoYYC23} since we find that preposition question before document can result in a performance boost when trying \textsc{ICLCite} method in the preliminary experiments. The concrete templates are shown in Table~\ref{tab:template}.

\section{Details about the Instruction}
\label{appendix:details_instructions}
The templates employed for generating preference data are detailed in Table~\ref{tab:positive_template} for positive instances, Table~\ref{tab:hallu_template} for statements exhibiting hallucination errors, Table~\ref{tab:syn_template} for statements with synthesis errors, and Table~\ref{tab:omission_template} for statements characterized by omission errors.
\section{Case Study}
\label{appendix:ea}
\begin{table*}[h!]
\begin{tcolorbox}
{\bf Question}\\ 
When did the rams go to st louis?

{\bf Documents}\\
Document [1](Title: History of St. Louis): 2011, with performances by Jay Leno and Aretha Franklin. In January 1995, Georgia Frontiere, the owner of the National Football League team known as the Los Angeles Rams (now St. Louis Rams), announced she would move that team to St. Louis. The team replaced the St. Louis Cardinals (now Arizona Cardinals), an NFL franchise that had moved to St. Louis in 1960 but departed for Arizona in 1988. \textcolor{red}{The Rams played their first game in their St. Louis stadium, the Edward Jones Dome, on October 22, 1996.} Starting in the early 1980s, more rehabilitation and construction projects began, some of

Document [2](Title: History of the St. Louis Rams): History of the St. Louis Rams The professional American football franchise now known as the Los Angeles Rams played in St. Louis, Missouri, as the St. Louis Rams from the 1995 through the 2015 seasons before relocating back to Los Angeles where the team had played from the 1946 season to the 1994 season. The Rams franchise relocated from Los Angeles to St. Louis in 1995, which had been without a National Football League (NFL) team since the Cardinals moved to Phoenix, Arizona in 1988. \textcolor{red}{The Rams’ first home game in St. Louis was at Busch Memorial Stadium against the}

Document [3](Title: History of the St. Louis Rams): History of the St. Louis Rams The professional American football franchise now known as the Los Angeles Rams played in St. Louis, Missouri, as the St. Louis Rams from the 1995 through the 2015 seasons before relocating back to Los Angeles where the team had played from the 1946 season to the 1994 season. The Rams franchise relocated from Los Angeles to St. Louis in 1995, which had been without a National Football League (NFL) team since the Cardinals moved to Phoenix, Arizona in 1988. \textcolor{red}{The Rams’ first home game in St. Louis was at Busch Memorial Stadium against the}

Document [4](Title: Los Angeles Rams): in 1980. After the 1994 NFL season, the Rams left California and moved east to St. Louis, Missouri. Five seasons after relocating, the team won Super Bowl XXXIV in a 23–16 victory over the Tennessee Titans. They appeared again in Super Bowl XXXVI, where they lost 20–17 to the New England Patriots. The Rams continued to play in Edward Jones Dome in St. Louis until the end of the 2015 NFL season, when the team filed notice with the NFL of its intent to pursue a relocation back to Los Angeles. The move was approved by a 30–2 margin at

Document [5](Title: 1994 Los Angeles Rams season): 1994 Los Angeles Rams season The 1994 Los Angeles Rams season was the franchise’s 57th year with the National Football League and the 49th and last season in the Greater Los Angeles Area until their 2016 relocation back to Los Angeles. After nearly 50 years in the Greater Los Angeles Area, owner Georgia Frontiere announced that the team would relocate to St. Louis, Missouri on January 15, 1995. While the owners initially rejected the move, permission was eventually granted therefore bringing an end to Southern California's first major professional sports franchise until 2016. The threat of relocation dominated talk about

\tcblower
{\bf Output}  \\
...Their first home game in St. Louis was at Busch Memorial Stadium against the Chicago Bears on October 22,1996 [1]....\\
\end{tcolorbox}
\caption{Sample containing fabrication error. In this sample, \textbf{Chicago Bears} does not appear in the reference documents.}
\label{tab:errors_hallu}
\end{table*} 

\begin{table*}[h!]
\begin{tcolorbox}
{\bf Question}\\ 
Who performed at the champions league final 2018?

{\bf Documents}\\
Document [1](Title: 2016 UEFA Champions League Final): worldwide via UEFA.com from 1 to 14 March 2016 in four price categories: €440, €320, €160 and €70. The remaining tickets were allocated to the local organising committee, UEFA and national associations, commercial partners and broadcasters, and to serve the corporate hospitality programme. American singer Alicia Keys performed in the opening ceremony prior to the match, the first time it has featured a live music performance. Italian tenor Andrea Bocelli performed the UEFA Champions League Anthem. The 2016 UEFA Women's Champions League Final was held two days prior, on 26 May 2016, at the Mapei Stadium – Città del Tricolore

Document [2](Title: 2018 UEFA Champions League Final): Lipa performed at the opening ceremony preceding the final. Jamaican rapper Sean Paul joined her as a special guest to perform their collaborative song, \"No Lie\". The 2018 UEFA Women's Champions League Final was held two days earlier, on 24 May 2018, at the Valeriy Lobanovskyi Dynamo Stadium between Wolfsburg and Lyon, Lyon emerging victorious 4–1. This was also the last time that the host city for the men's Champions League final was also automatically assigned the Women's Champions League final. The annual UEFA Champions Festival was held between 24–27 May 2018 at the Kiev city centre. In late May,

Document [3](Title: UEFA Champions League Anthem): the two teams are lined up, as well as at the beginning and end of television broadcasts of the matches. Special vocal versions have been performed live at the Champions League Final with lyrics in other languages, changing over to the host country's language for the chorus. These versions were performed by Andrea Bocelli (Italian) (Rome 2009, Milan 2016 and Cardiff 2017), Juan Diego Flores (Spanish) (Madrid 2010), All Angels (Wembley 2011), Jonas Kaufmann and David Garrett (Munich 2012), Mariza (Lisbon 2014, unlike the previous final performers, Mariza sang the main lyric of the anthem), and Nina Maria Fischer and

Document [4](Title: 2018 UEFA Champions League Final): 2018 UEFA Champions League Final The 2018 UEFA Champions League Final was the final match of the 2017–18 UEFA Champions League, the 63rd season of Europe's premier club football tournament organised by UEFA, and the 26th season since it was renamed from the European Cup to the UEFA Champions League. It was played at the NSC Olimpiyskiy Stadium in Kiev, Ukraine on 26 May 2018, between Spanish side and defending champions Real Madrid, who had won the competition in each of the last two seasons, and English side Liverpool. \textcolor{red}{Real Madrid won the final 3–1 for their third consecutive and}

Document [5](Title: 2017 UEFA Champions League Final): 2017 UEFA Champions League Final The 2017 UEFA Champions League Final was the final match of the 2016–17 UEFA Champions League, the 62nd season of Europe's premier club football tournament organised by UEFA, and the 25th season since it was renamed from the European Champion Clubs' Cup to the UEFA Champions League. It was played at the Millennium Stadium in Cardiff, Wales on 3 June 2017, between Italian side Juventus and Spanish side and title holders Real Madrid, in a repeat of the 1998 final. \textcolor{red}{Real Madrid won the match 4–1 to secure their 12th title in this competition.} With
\tcblower
{\bf Output}  \\
...Real Madrid won the final 3-1 for their third consecutive and overall 12th title in this competition[4]...\\
\end{tcolorbox}
\caption{Sample containing synthesis error. In this sample, document 4 and document 5 are mistakenly intermingled.}
\label{tab:errors_synthesis}
\end{table*}

\begin{table*}[h!]
\begin{tcolorbox}
{\bf Question}\\ 
When was the 13th amendment ratified by the states?

{\bf Documents}\\
Document [1](Title: Thirteenth Amendment to the United States Constitution): ratification did not imply federal power to legislate on the status of former slaves. During the first week of December, North Carolina and Georgia gave the amendment the final votes needed for it to become part of the Constitution. The Thirteenth Amendment became part of the Constitution on December 6, 1865, based on the following ratifications: Having been ratified by the legislatures of three-fourths of the several states (27 of the 36 states, including those that had been \"in rebellion\"), Secretary of State Seward, on December 18, 1865, certified that the Thirteenth Amendment had become valid, to all intents and

Document [2](Title: Thirteenth Amendment to the United States Constitution): Thirteenth Amendment to the United States Constitution The Thirteenth Amendment (Amendment XIII) to the United States Constitution abolished slavery and involuntary servitude, except as punishment for a crime. In Congress, it was passed by the Senate on April 8, 1864, and by the House on January 31, 1865. The amendment was ratified by the required number of states on December 6, 1865. On December 18, 1865, Secretary of State William H. Seward proclaimed its adoption. It was the first of the three Reconstruction Amendments adopted following the American Civil War. Since the American Revolution, states had divided into states that

Document [3](Title: Emancipation Proclamation): Winning re-election, Lincoln pressed the lame duck 38th Congress to pass the proposed amendment immediately rather than wait for the incoming 39th Congress to convene. In January 1865, Congress sent to the state legislatures for ratification what became the Thirteenth Amendment, banning slavery in all U.S. states and territories. The amendment was ratified by the legislatures of enough states by December 6, 1865, and proclaimed 12 days later. There were about 40,000 slaves in Kentucky and 1,000 in Delaware who were liberated then. As the years went on and American life continued to be deeply unfair towards blacks, cynicism towards

Document [4](Title: Thirteenth Amendment to the United States Constitution): Enforcement, and Contemporary Implications Thirteenth Amendment to the United States Constitution The Thirteenth Amendment (Amendment XIII) to the United States Constitution abolished slavery and involuntary servitude, except as punishment for a crime. In Congress, it was passed by the Senate on April 8, 1864, and by the House on January 31, 1865. \textcolor{red}{The amendment was ratified by the required number of states on December 6, 1865.} On December 18, 1865, Secretary of State William H. Seward proclaimed its adoption. It was the first of the three Reconstruction Amendments adopted following the American Civil War. Since the American Revolution, states had

Document [5](Title: Thirteenth Amendment to the United States Constitution): \textcolor{red}{joint resolution was sent to Lincoln for his signature.} Under the usual signatures of the Speaker of the House and the President of the Senate, \textcolor{red}{President Lincoln wrote the word \"Approved\" and added his signature to the joint resolution on February 1, 1865. On February 7, Congress passed a resolution affirming that the Presidential signature was unnecessary.}  The Thirteenth Amendment is the only ratified amendment signed by a President, although James Buchanan had signed the Corwin Amendment that the 36th Congress had adopted and sent to the states in March 1861. When the Thirteenth Amendment was submitted to the states

\tcblower
{\bf Output}  \\
...The amendment was ratified by the required number of states on December 6,1865, and President Lincoln signed the joint resolution on February 1,1865, after it was passed by Congress on February 7,1865 [5]...\\
\end{tcolorbox}
\caption{Sample containing omission error. In this sample, document 4 is not attributed.}
\label{tab:errors_omission}
\end{table*} 

In this section, we perform a detailed case study and demonstrate several examples of each type of error we defined. As shown in Table~\ref{tab:errors_hallu}, we classify it as a fabrication error since it uses an undefined entity. In Table~\ref{tab:errors_synthesis}, we classify it as a synthesis error since it mixes up facts from document 4 and document 5, which results in a factual error. In Table~\ref{tab:errors_omission}, we classify it as a omission error since it used facts from document 4 and document 5, but document 4 is not attributed.

\section{Related Works of Retrieval Augmentation of LLMs}

Retrieval augmentation has emerged as a prominent technique aimed at enhancing the accuracy and veracity of LLMs~\cite{gao2023retrieval,DBLP:conf/acm/AsaiMZC23}. Specifically, \citet{DBLP:journals/corr/abs-2312-09075} couples LLMs with long-term and short-term memories, resulting in improved claim and citation generation. Meanwhile, in order to effectively incorporate external knowledge into LLMs, SearChain proposes a global reasoning chain strategy that facilitates retrieval augmentation generation at each node within the chain~\cite{DBLP:journals/corr/abs-2304-14732}. In another line of research, the self-reflection is leveraged for retrieval verification during the retrieval-augmented generation process~\cite{DBLP:journals/corr/abs-2311-07838,DBLP:journals/corr/abs-2310-11511}. Despite these advancements, prior studies have not adequately addressed the issue of attribution hallucination~\cite{zuccon2023chatgpt}. In contrast, we focus on making the model better answer the query and align with the reference.

\begin{figure*}[htb]
    \centering
   \resizebox{\linewidth}{!}{
    \includegraphics[]{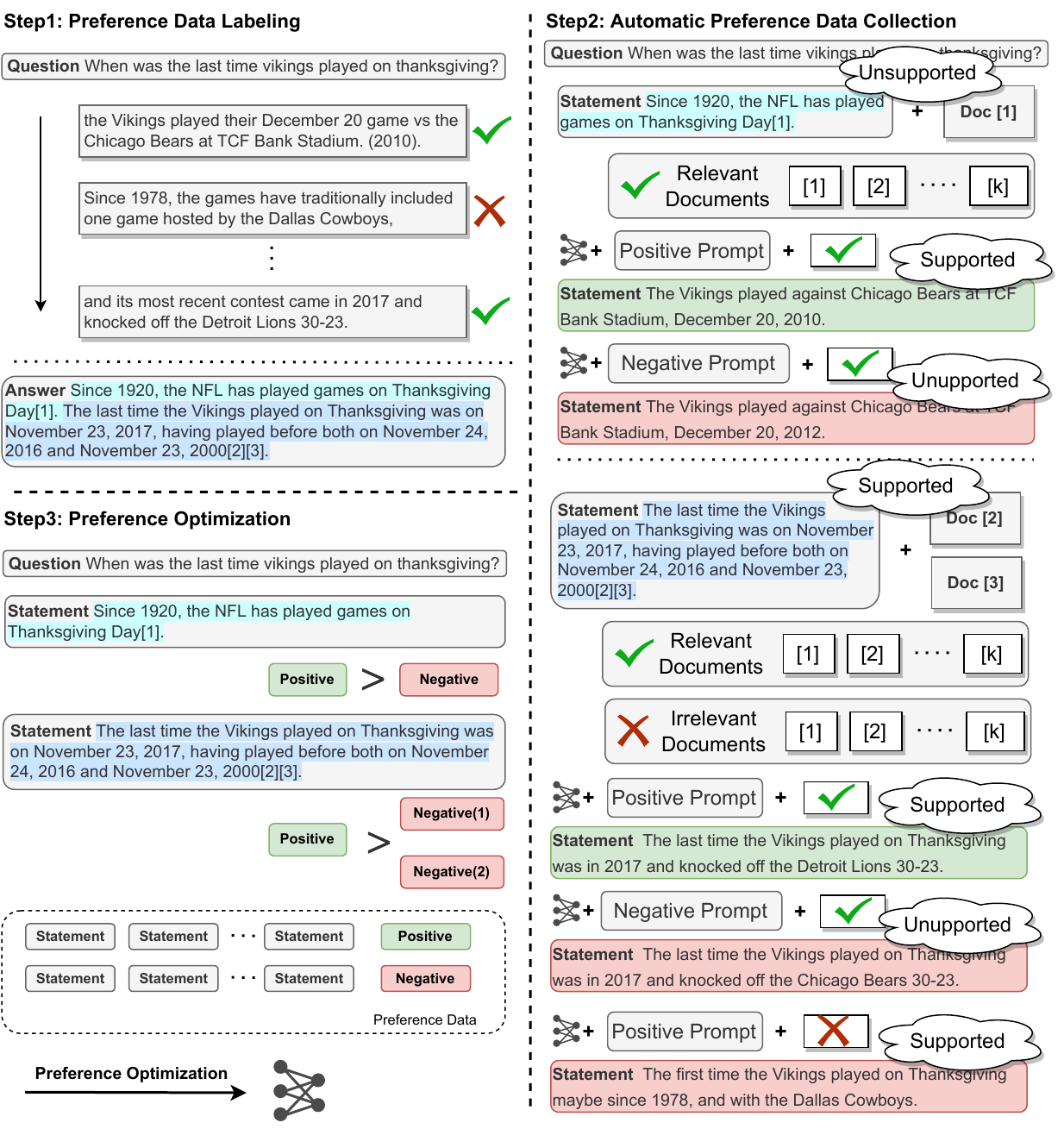}
    }
    \caption{
    The overall framework of the APO. 
    }
    
    \vspace{-4mm}
    \label{fig:framework}
\end{figure*}

\begin{algorithm*}
\caption{Automatic preference data collection algorithm}
\begin{algorithmic}[1]
\State \textbf{Input} Input $P_{init}$, Critic $M_c$, Generator $M_g$, Retriever $R$, Error instructions $\mathcal{E}$
\State \textbf{Output} Output augmented preference dataset $P_{syn}$
\State $P_{tmp}$ = \{\}
\For{$(q, D, \mathcal{S}, \mathcal{C}, \mathcal{L}_{rel}, \mathcal{L}_{sup})$ in $P_{init}$}
    \For {each statement $s_i$ in $\mathcal{S}$}
        \If{the referenced passages $\mathcal{C}_i$ are all \textit{relevant} to $q$}
            \State add $(q, D, \mathcal{S}_{:i}, \mathcal{C}_{:i}, \mathcal{L}_{rel}, \mathcal{L}_{sup}^{:i,:})$ to $P_{tmp}$
        \EndIf
    \EndFor
\EndFor

\For{$(q, D, \mathcal{S}_{:i}, \mathcal{C}_{:i}, \mathcal{L}_{rel}, \mathcal{L}_{sup}^{:i,:})$in $P_{tmp}$}
    \State Retrieve top-$m$ passages $D_{ir}$ using retriever $R$ given $q$.
    \State Subsequently delete passage $d$ from $D_{ir}$, if $d_i$ is predicted as \textit{relevant} to $q$ using critic $M_c$.
    \For{each statement $s_i$ in $\mathcal{S}$ and its relative attributed passages $\mathcal{C}_i$}
        \If{$s_i$ is supported by $\mathcal{C}_i$}
            \State Predicts \textit{supported} for $s_i$ using critic $M_c$ given $q, d_{i}, s_i$, where $d_{i}\in D$
            \State Add \textit{supported} passages to the relative attributed passages $\mathcal{C}_i$ of statement $s_i$.
            \State Generate $s_i^{s\wedge r}$ using $M_g$ given $q$, $s_{:i-1}$, new $\mathcal{C}_i$.
            \State Generate $s_i^{s\wedge  \tilde{r}}$ using $M_g$ given $q$,$s_{:i-1}$, $D_{ir}$.
            \State Generate $s_i^{ \tilde{s} \wedge r}$ using $M_g$ given $q$, $s_{:i-1}$, new $\mathcal{C}_i$ and pre-defined error type $e$.
             \State add $(q,  s_i^{s\wedge r}, s_i^{ \tilde{s} \wedge r}, D$) and  $(q,  s_i^{s\wedge r}, s_i^{s\wedge  \tilde{r}}, D$) to $P_{syn}$.
        \Else
            \State Generate $s_i^{s\wedge r}$ using $M_g$ given $q$, $s_{:i-1}$, new $\mathcal{C}_i$.
            \State Generate $s_i^{ \tilde{s}\wedge r}$ using $M_g$ given $q$, $s_{:i-1}$, $D -$ new $\mathcal{C}_i$ and pre-defined error type $e$.
            \State add $(q,  s_i^{s\wedge r}, s_i^{ \tilde{s} \wedge r}, D$) to $P_{syn}$.
        \EndIf
        
    \EndFor
\EndFor 

\end{algorithmic}
\label{preference_data_collection_step2}
\end{algorithm*}

\end{document}